# Classification of 12-Lead ECG Signals with Bi-directional LSTM Network

Ahmed Mostayed, Junye Luo, Xingliang Shu, William Wee

Multimedia and Augmented Reality Lab, Department of Electrical Engineering and Computing Systems, University of Cincinnati

## Abstract

We propose a recurrent neural network classifier to detect pathologies in 12-lead ECG signals and train and validate the classifier with the Chinese physiological signal challenge dataset (http://www.icbeb.org/Challenge.html). The recurrent neural network consists of two bi-directional LSTM layers and can train on arbitrary-length ECG signals. Our best trained model achieved an average F1 score of 74.15% on the validation set.

**Correspondence:**

Ahmed Mostayed

Email: mostayad@mail.uc.edu

Engineering Research Center Room 817

University of Cincinnati

2600 Clifton Avenue, Cincinnati, OH 45220, USA

# 1. Introduction

In this paper, we developed an algorithm to detect rhythm/morphology abnormalities from 12-lead ECG signals. The detection task was a part of 2018 China physiological signal challenge, an initiative to encourage development of open-source machine learning algorithms to automatically detect 8 abnormalities of 4 broad classes – atrial fibrillation (AF), blocks, premature contraction, and ST-segment abnormalities. We designed an end-to-end shallow 3-layer recurrent neural network (RNN) [1] model (two hidden recurrent layers followed by a feed-forward classification layer) that was trained on ECG segments (typically 6 – 10 seconds long) extracted from ECG signals (up to 60 seconds long). In test time, the prediction of each extracted segments was combined to predict the final class (one of nine, including normal) of the ECG sequence.

Automatic detection of pathologies or mental states from physiological time-series like EEG or ECG is a relatively new research field, dating back to the early 90s. Due to the variable data length of ECG signals, RNNs, in one form or other, has been the architecture of choice for such tasks. Lipton et al. [2] proposed a LSTM [3] network for multi-label classification task which treats time-series of electronic health records as a sequence of observations (corresponding to each time sample). They defined a loss function which consists of a weighted sum of the loss at the final sequence step and the average loss at all the previous steps. Bashivan et al. [4] employed a convolutional recurrent neural network (CRNN) architecture to classify mental states from EEG signals. The CRNN is an end-to-end model which consists of one or two recurrent layers on top of a series of convolutional layers. Following this approach, Zihlmann et al. [5] developed a CRNN model for AF classification on the Physionet/CinC challenge 2017 [6] dataset. Their approach involves a series of 2D convolutions of time-frequency representation of the ECG signal, obtained via spectrogram calculation, followed by flattening (to create a sequence), a recurrent layer, and a classification layer. Hwang et al. [7] took an approach like [5] for classifying mental states from ECG signals. Instead of calculating the spectrogram, they opted to perform 1D convolutions for raw ECG data. Rajkapur et al. [8] treated ECG arrhythmia detection as a sequence-to-sequence learning, where each ECG signal is treated as a sequence of fixed-length short segments and the label for each signal is treated as a sequence of annotation for each segment. They developed a 34-layer deep CNN architecture consisting of residual blocks [9]. The paper claims cardiologist-level accuracy for 13 arrhythmia types for a model trained on a dataset of 29, 163 individuals.

Our RNN model is most closely related to the model of Lipton et al. [2]. Like them we have employed a network with 2 hidden recurrent layers. However, there are two distinctions. First, our LSTM cells are bi-directional [10]. Second, we do not make use of the outputs of the intermediate time steps to calculate the loss function. We trained and validated our model on training and validation sets constructed from the 6, 877 individuals in the China physiological signal dataset. Our model achieved an overall F1 score of 74.15% on the validation set for the classification task of 9 classes (normal plus 8 abnormalities). An earlier version of the model was submitted to the challenge to evaluate on an unseen test dataset which yielded F1 score of 65.8%. At the time of writing the current model was not submitted for evaluation.

## 2. Materials and Methods

### 2.1. Problem Statement

Given a collection of 12-lead ECG signals for 9 classes, including the normal heart condition and 8 abnormal conditions, we formulate the abnormality detection task as a pattern classification problem. A 12-lead ECG signal with $T$ time samples can be presented as a sequence of 12-dimensional vectors of length $T$. Formally, given a sequence $X = \{\boldsymbol{x}[0], \boldsymbol{x}[1], \boldsymbol{x}[2], \cdots\cdots \boldsymbol{x}[T]\}$, a classifier is trained to learn the probabilities of 9 classes (one normal and 8 abnormal):

$$\hat{y} = p_L(L = l|X) \qquad l = 1,2,\cdots\cdots,9 \qquad (1)$$

where $T$ and $L$ are the length and label of the sequence respectively, and $\boldsymbol{x}[n]\epsilon\mathbb{R}^{12\times1}$ is the input vector at time $n$, and $\hat{y}\epsilon\mathbb{R}^{9\times1}$ is the probability outputs.

### 2.2. Model Architecture

We used a RNN with 2 hidden recurrent layers with 100 recurrent cells each and 1 fully-connected classification (output) layer. A schematic of the network architecture is given in Figure 1. The recurrent cells are bi-directional LSTM [10] cells. Each bi-directional cell consists of a forward and a backward stream and at each time step the outputs of both streams are combined (element-wise multiplication for the first hidden layer to obtain a 100-dimensional representation). The output of the second hidden layer for both streams at the last time step is concatenated and fed to the classification layer. The classification layer uses softmax activation function to obtain the final 9-dimensional output. To regularize the network, we also applied dropout (with drop probability of 50%) [11] to the activations of each hidden layer. We did not use any batch normalization.

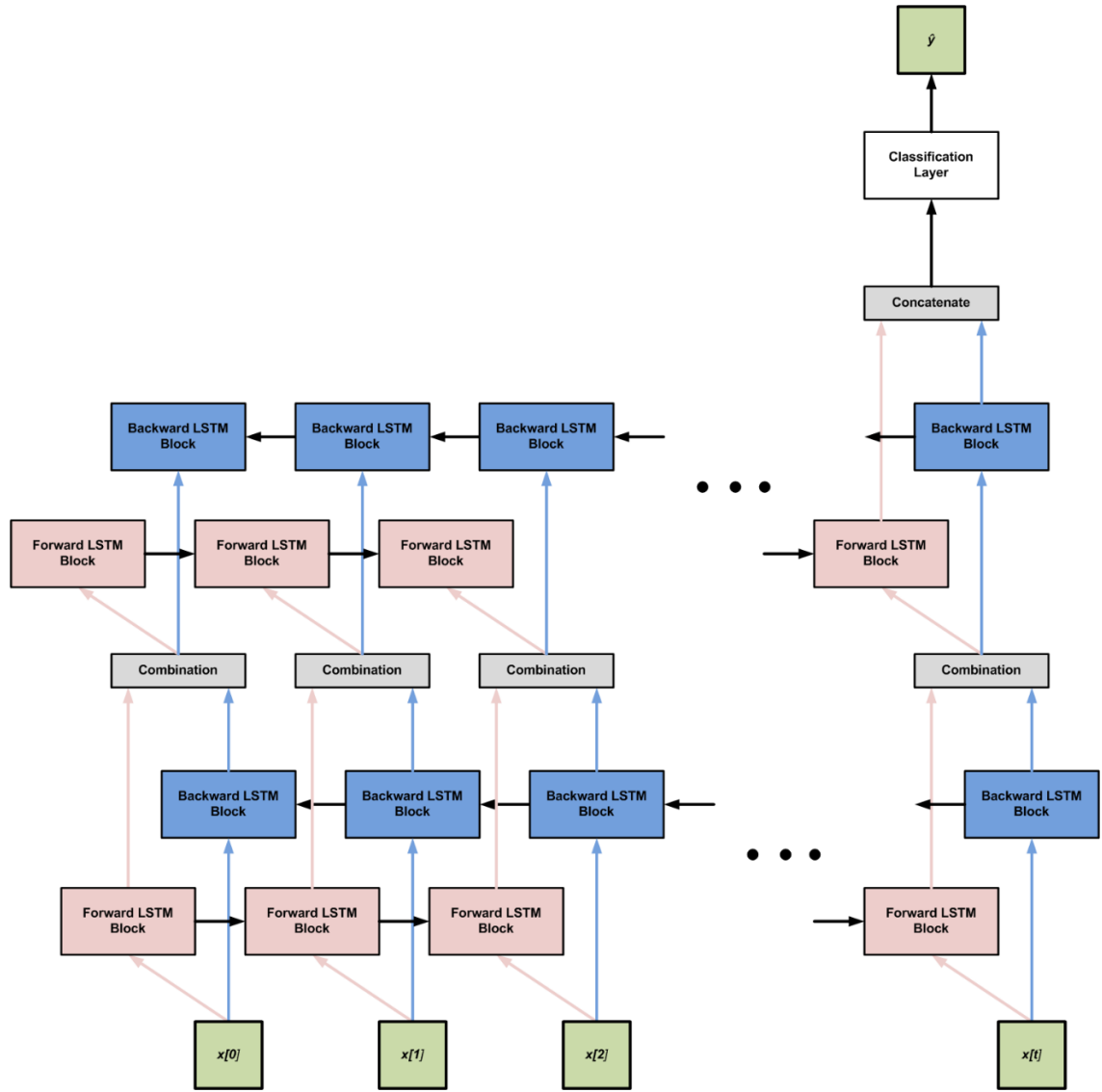

Figure 1. The architecture of the RNN model. The network has 2 bi-directional LSTM hidden layers, followed by a fully-connected layer with softmax activation.

## 2.3. Data Preparation

The ECG recordings in the Chinese physiological dataset varies between 6 seconds to 60 seconds in duration, with a mean duration of 15.79 seconds. Given the high sampling rate (500 Hz) of the data acquisition, the number of time samples could be very high for many individuals. Although LSTM are known to be very robust for long time dependencies, such long sequences could still be daunting for the network. Therefore, we trained our network with small segments (children) extracted from a given ECG signal (parent). A typical segment consisted of 4 cardiac beats. Each segment from the training set was appropriately annotated with the right class label. For AF, PAC or PVC the pathology only occurs at certain cardiac cycles of the parent ECG and children which include those are labelled as the pathology and the rest are labelled as normal. Segments coming from the other 6 classes are labelled as the class label of their parents.

We developed an automated algorithm to extract the segments and to annotate them. The algorithm is based on the QRS signal detection using stationary wavelet transform [12]. QRS detection of ECG signal is a matured research area and hence, we do not delve into the details of the algorithm in this paper. However, readers are directed to the articles referenced in [12 - 15], as our method is an amalgamation of the techniques described in them.

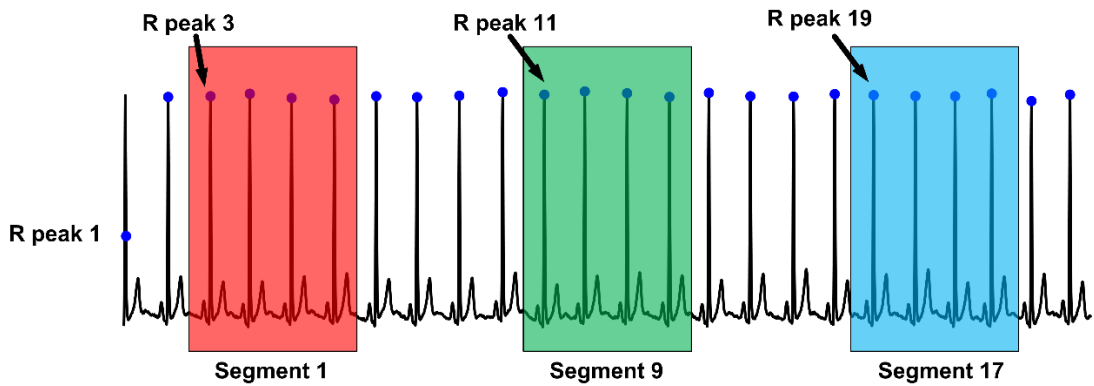

Figure 2. An ECG signal (A0016) and corresponding R wave peaks detected by the QRS algorithm shown as blue dots. The algorithm detected 24 peaks. 17 segments were automatically extracted (each consisting of 4 R wave peaks) from the data. Note that segments corresponding to the first and last 2 peaks were excluded as those segments often exhibit unpredictable shapes (probably caused by sudden attachment and detachment of the electrodes)

We further preprocessed the segmented ECG signals to correct baseline wander and remove high frequency noise. We used a Butterworth high-pass filter with cut-off frequency 1 Hz to remove the baseline wander. High frequency noise was removed using wavelet transform-based shrinkage methods [16].

### 2.4. Training

The network was trained by minimizing the cross-entropy loss function:

$$\mathcal{L}(X, L) = \sum_{l=1}^{9} \log p_L(L = l | X) \qquad (2)$$

The weights were initialized using the He initialization scheme [17]. We used Adam optimizer [18] with default parameters and an initial learning rate of 0.001.

### 2.5. Validation

For validation each parent ECG signal is divided into children segments using the technique described in section 2.3 and their class labels are predicted by the model. The class label of the parent ECG is then determined as the class of majority of the children.

## 3. Data

The China physiological challenge dataset consists of 6,877 publicly available 12-lead ECG recordings lasting from 6 seconds to 60 seconds. The data includes one normal class and 8 abnormal classes. There is also a test set of 2,954 recordings which is currently not available in the public domain. The ECG recordings were acquired with a sampling rate of 500 samples per second. The recordings are class labelled for the entire length of the recordings and not for specific portions of the data. Some recordings have more than one (up to 3) annotated classes, without any mention of the location or duration of the individual abnormalities. Table 1 shows the distribution of the publicly available dataset over 9 classes.

Table 1. Dataset distribution over 9 classes.

| Type | Class | Number of Records |
|---|---|---|
| | Normal | 918 |
| | Atrial fibrillation (AF) | 1098 |
| Block | First-degree atrioventricular block (I-AVB) | 704 |
| | Left bundle branch block (LBBB) | 207 |
| | Right bundle branch block (RBBB) | 1695 |
| Premature contraction | Premature atrial contraction (PAC) | 556 |
| | Premature ventricular contraction (PVC) | 672 |
| ST-segment abnormalities | ST-segment depression (STD) | 825 |
| | ST-segment elevated (STE) | 202 |

### 3.1. Training

To train our RNN we randomly split the entire dataset of 6,877 recordings into training and validation sets with a 90/10 split ratio. That gives us 6,190 recordings for training and the rest for validation. Following the steps described in section 2.3 we extracted 87, 585 segments (14 segments per record on an average) from the training set, each appropriately labelled. It not only allows to keep the complexity of the network moderate, but also serves as a data augmentation step. As an additional data augmentation, we also included the raw

ECG segments (without baseline correction and noise removal) in the training data. Moreover, we down-sampled the signal segments to 70 samples per second to further reduce the computational burden.

## 3.2. Validation

From the 687 validation recordings, 9,829 segments are extracted. Unlike the training segments, the validation segments were not pre-processed to remove baseline wander or noise. Individual predictions obtained from the raw (but down-sampled) segments were combined (as described in section 2.5) to make the final prediction for an ECG recording.

# 4. Results

## 4.1. Evaluation Metrics

The accuracy of the model was evaluated on the validation set by calculating the per-class F1 score and the average F1 score. The per-class F1 score is calculated as,

$$F_{1,i} = \frac{2 \times N_{ii}}{\sum_{j=1}^{9} (N_{ij} + N_{ji})} \tag{3}$$

Where, $N_{ij}$ is the number of validation examples of class $i$ predicted as class $j$. We used 2 different formulas to calculate the average F1 score. First by taking the simple mean of the per-class F1 score,

$$F_1 = \frac{\sum_{i=1}^{9} F_{1,i}}{9} \tag{4}$$

And second, by calculating the class-frequency weighted mean,

$$F_1 = \sum_{i=1}^{9} \frac{N_i}{N} F_{1,i} \tag{5}$$

Where, $N$ is the total number of validation records and $N_i$ is the total of number of validation records belonging to class $i$.

## 4.2. Model Performance on Validation

Table 1 shows the F1 scores of the model prediction for the validation dataset. The model performs the best on RBBB and the worst on STE.

Table 2. F1 scores of individual classes on the validation set.

| Class | F1 Score |
|-------|----------|
| Normal | 0.7388 |
| AF | 0.7681 |
| I-AVB | 0.7419 |
| LBBB | 0.7058 |
| RBBB | 0.8215 |
| PAC | 0.5909 |
| PVC | 0.8070 |
| STD | 0.6582 |
| STE | 0.2941 |

Table 3 shows the average F1 score of the model prediction.

Table 3. Average F1 score on the validation set.

| Simple Mean | Weighted Mean |
|-------------|---------------|
| 0.7415 | 0.7388 |

Figure 3 shows the confusion matrix of the model prediction on the validation set.

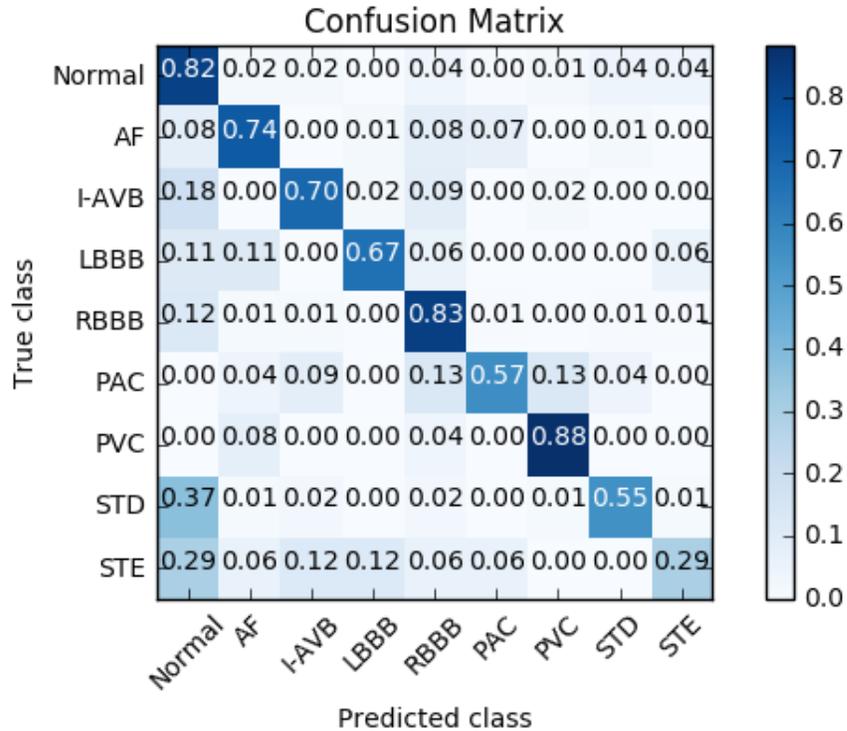

Figure 3. Confusion matrix of the model prediction for the validation set. The value shown on the $(i, j)$ cell represents the quantity $\frac{N_{ij}}{\sum_{j=1}^{9} N_{ij}}$.

## 5. Discussions

The off-diagonal entries of the confusion matrix in Figure 3 shows the proportions of instances of each class in the validation set identified as other classes. The first row of the confusion matrix indicates that roughly 18% of the normal ECG were misclassified as abnormal. Our visual inspection revealed that in many cases signals labelled as normal have patterns associated with a pathology, forcing the RNN model to make a wrong prediction (see Figure 4). At times poor signal-to-noise ratio of the signal cam cause misclassification, as for A5909. The confusion matrix also indicates that 22% of LBBB examples are identified as either normal or AF. Figure 5 shows two such examples. The QRS complex in the left example in Figure 5 resembles a normal ECG. On the other hand, the signal on the right exhibits R-R interval variations generally associated with AF.

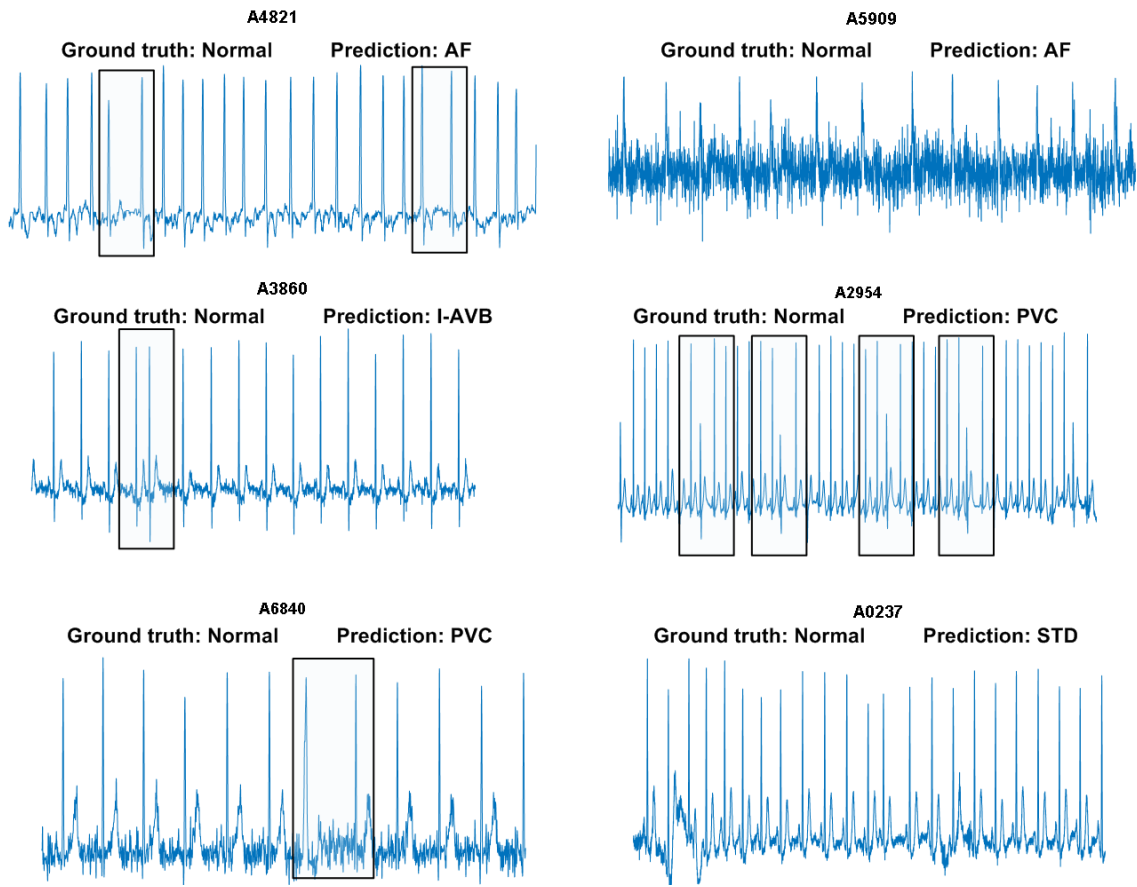

Figure 4. Some examples of normal ECG signals classified as abnormal. Visual inspection reveals that these signals are misclassified either due to presence of excessive noise (A5909) or presence of patterns resembling pathology (marked with the black rectangles).

Our model performs poorly on the ST-segment abnormalities. Almost 30% of examples belonging these two classes are misclassified as normal despite having clear visual evidence of pathology. Some examples are shown in Figure 6.

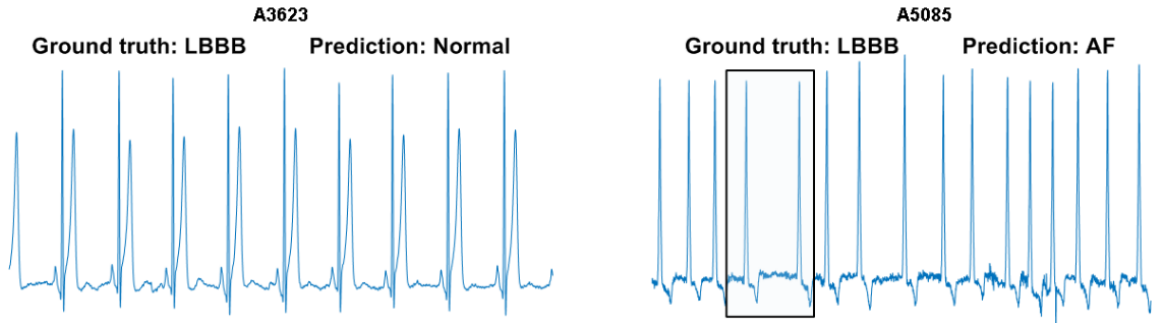

Figure 5. Examples of wrongly classified LBBB signals. A3626 visually appears like a normal ECG signal. A5085 is most likely classified as AF due to the pattern appearing on the portion of the signal within the rectangular window.

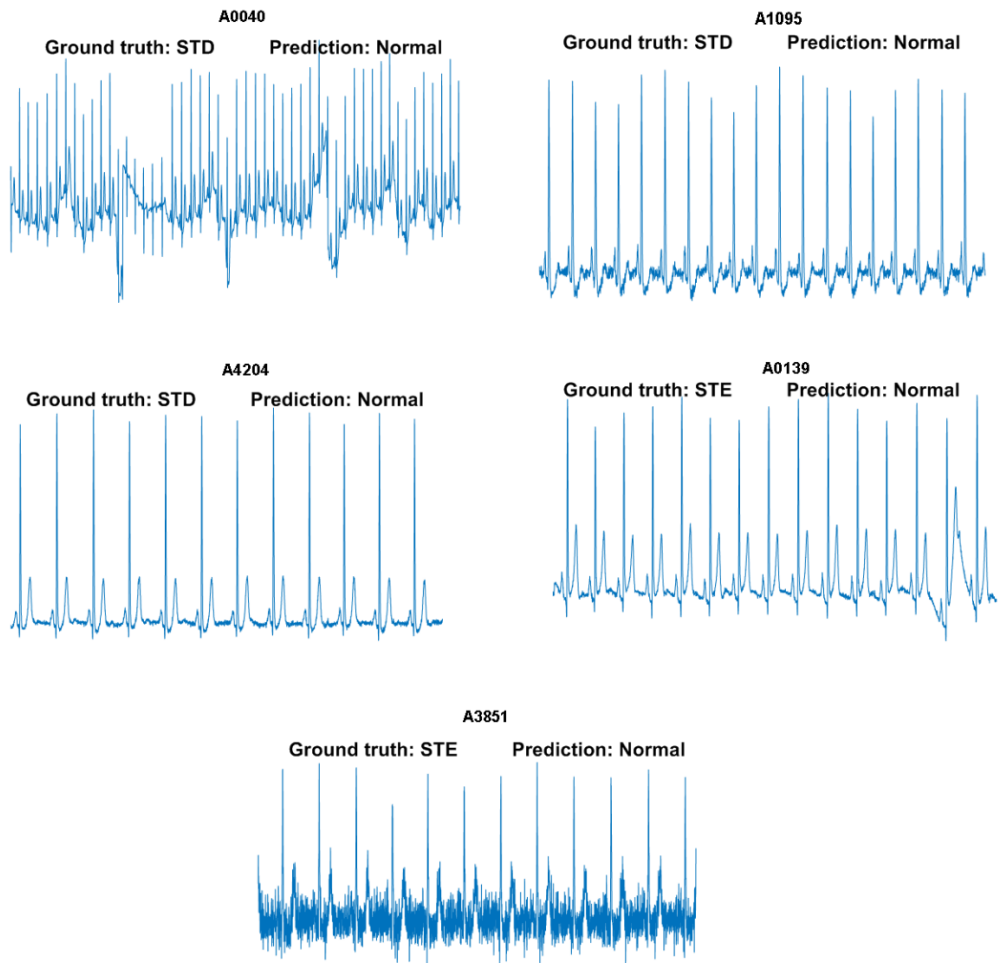

Figure 6. Examples of ST-segment abnormalities classified as normal.

## 6.  Conclusion

We developed an RNN model to detect 8 different abnormalities from 12-lead ECGs. The model achieved an average F1 score of 0.7415 on a validation set of 687 recordings. Although we did not submit our model to test against the unseen test dataset (as of this writing), we are confident of achieving similar performance. We recognize our model has performed considerably poorly for ST-segment abnormalities compared to the top performed models submitted to the challenge.  However, our model performed on par with those models for the other 7 classes. For future work, we intend to train a CRNN [4] type model that will learn fixed-length features on small segments of the ECG signal (QRS, ST-segment, etc.) via an unsupervised CNN, and the learned features will then be sequentially fed to a RNN to make the class prediction.